# Transliteration of Foreign Words in Burmese

**Chenchen Ding**
ASTREC, NICT
3-5 Hikaridai, Seika-cho, Soraku-gun, Kyoto, 619-0289, Japan
chenchen.ding@nict.go.jp

## Abstract

This manuscript provides general descriptions on transliteration of foreign words in the Burmese language. Phenomena caused by phonetic and orthographic issues are discussed. Based on this work, we expect to gradually establish prescriptive guidelines to normalize the transliteration on modern words in Burmese.

## 1 Introduction

In the process of language modernization, the Burmese language has borrowed lots of foreign words for new concepts. Traditional loanwords were mainly from Sanskrit and Pali. Modern terminologies are mainly borrowed from English. A huge number of non-native words in popular culture also enter daily communication.

Due to the features of Burmese, the spelling of foreign words has a certain inconsistency. Some major rules can be observed, while there is still a large margin of varieties. This manuscript is a descriptive work on issues of transliteration of foreign words in Burmese.

We expect to gradually establish relatively prescriptive guidelines for the foreign words transliteration in Burmese based on this work. This work will also contribute to textual data normalization for the application of language processing techniques on Burmese.

In this manuscript, we mainly focus on the transliteration from English to Burmese, as English is the most common source of the borrowed words in Burmese. Furthermore, general Latin letter spelt foreign words, even non-native English, are commonly treated as if they are English in the transliteration. An appendix on the transliteration from Chinese to Burmese based on the Pinyin system is also provided.

## 2 Background and Related Work

As to the phonology of source languages, Hammond (1999) and Duanmu (2007) have provided comprehensive descriptions on English and Chinese, respectively. An introduction of phonology, orthography, morphology, and syntax of Burmese can be referred to Okano (2007). As to the related research in the field of natural language processing, Ding et al. (2018) handled the Romanization of Burmese names and provided a grapheme-aligned dataset. Aye Myat Mon et al. (2020) provided a large dataset of transliterated names in Burmese. This work is generally based on the two datasets for the investigation.

## 3 Notation

As to the notation in this manuscript, phonemes are put into a pair of "//"; graphemes are put into a pair of "<>". Symbols of International Phonetic Alphabet (IPA, 1999) are generally used. For Burmese consonants, /ʃ/, /tʃ/, and /dʒ/ are used for the palatalized consonants and /θ/ used for the dental /t/ (Okano, 2007). These notations are shared with those used for English. For Burmese rhymes, the glottal ending is marked by "ʔ", and the nasalization by "ℵ".

Since the Burmese script is an abugida system, there is no letter for standalone consonants at onset but always an inherent vowel is attached. We use a "-" after a Burmese consonant letter to mark the deletion of the inherent vowel. All the graphemes to alternate the inherent vowel will begin with a "+". The inherent vowel will be represented by a bare "+" when necessary. The following is an example by this notation to decomposition the instance <ALEXANDER> → <အလက်ဇန္ဒား> into alignment of graphemes.

| | | |
|---|---|---|
| <> | → | <အ-> |
| <A> | → | <+> |
| <L> | → | <လ-> |
| <EX> | → | <+က်ဆ-> |
| <AND> | → | <+န္ဒ-> |
| <A> | → | <+တ၀ေး> |

## 4 Analysis

### 4.1 Data and Examples

The dataset by Aye Myat Mon et al. (2020) were collected from textual data on the Internet with considerable noise. With the overall tendencies by this dataset, we refer to two sources for specific examples: 1) the SEAlang digitalized Burmese-English Dictionary edited by Burmese Language Commission (MLC)[1] and 2) the contents from the Burmese version of Wikipedia[2]. Examples from the two sources are marked as [M] and [W], respectively.

### 4.2 Phonotactic Issues

The structure of Burmese syllables has stricter restrictions than that of English. Generally, the onset of an English syllable will be completely transliterated into Burmese by one or multiple consonant letters; the nucleus and sonorant consonants in coda (if any) will be transliterated as an integrated rhyme; obstruent consonants in coda are usually dropped, turning to a glottal ending or a creaky tone. Further irregular spellings may be used to indicate the dropped consonants in coda.

### 4.3 Simple Onset

Burmese has abundant consonants appearing at the onset of a syllable, which overlap to those in English to a large extent. Some strong phoneme-to-grapheme mapping used for transliteration can be summarized as follows: /p/ to <ပ->, /t/ to <တ->, /k/ to <က->, /b/ to <ဘ->, /d/ to <ဒ->, /g/ to <ဂ->, /z/ to <ဇ->, /dʒ/ to <ဂျ->, /m/ to <မ->, /n/ to <န->, /l/ to <လ->, /j/ to <ယ->, /w/ to <ဝ->, and /h/ to <ဟ->.

The aspiration of obstruents is not a distinguishing feature in English. Burmese letters for unaspirated voiceless obstruents are generally used. Meanwhile, there are two common mappings by aspiration letters/graphemes: /tʃ/ to <ချ-> and /ʃ/ to <ရှ->.

Aspiration letters are also used for absent phonemes in Burmese, i.e., /f/ to <ဖ-> (native /pʰ/).

The distinction between /s/ and /sʰ/ are disappearing in Burmese, so <စ-> and <ဆ-> are competitive for /s/. However, <စ> is more used in onset clusters such as <စတ-> for /st/.

The /r/ has disappeared in Burmese while the original letter "ရ" is kept by the orthography. So, <ရ-> for /r/ is common, although the use of <ရ-> and <လ-> are exchangeable to a certain extent.

<ဗ-> is common for the non-native phoneme /v/, which can distinguish from <ဘ-> for /b/.

The less common /ʒ/ may be treated as /dʒ/ or may follow some grapheme-based mapping.

The abovementioned mappings are from phonemes to graphemes, say, transcription rather than transliteration. Here is an example on the treatment of the letter "C", where /s/ to <ဆ-> and /k/ to <က-> are applied.

<CIRCUS> → <ဆပ်ကပ်> [M]

More often, the mapping will not go to the extent of phonemes, but conducted at the grapheme level, i.e., literally transliteration.

<TH> to <သ-> is a strong mapping between graphemes regardless of the phonemes. We have the following examples, where the <TH> is for /ð/ and /θ/ respectively.

<LOGARITHM> → <လော်ဂရစ်သမ်> [M]

<THEORY> → <သီအိုရီ> [M]

Sometimes strong grapheme-to-grapheme mappings override phoneme-to-grapheme mappings.

As an example, <J> to <ဇ-> is common in some stable borrowed words.

<JANUARY> → <ဇန်နဝါရီ> [M]

<JULY> → <ဇူလိုင်> [M]

<JUNE> → <ဇွန်> [M]

In these examples, <J> to <ဇ-> overrides /dʒ/ to <ဂျ->, but not in the following examples.

<JOURNAL> → <ဂျာနယ်> [M]

<JURY> → <ဂျူရီ> [M]

Moreover, the etymology may be involved.

<JESUS> → <ယေရှု> [M]

In this example, <J> to <ယ-> and <S> to <ရှ-> may be based on Biblical Hebrew. The following example is about a Spanish name.

<JUAN> → <ဝမ်> [W]

Here an underlying chain of <JU> → /xw/ → /hw/ → /w/ to <ဝ-> can be considered behind the surface <JU> to <ဝ->.

## 4.4    Onset Cluster

Burmese does not allow too complex onset clusters. Generally, the consonants in onset clusters from English will be represented by a sequence of basic consonant letters, where the final one will be further modified for the following nucleus.

The mappings follow those in the simple onset. In cluster ended with /r/, <ရ-> will be used without confusion of <ဝ->.

Special transliteration for clusters can be observed. As an example, <CHR> for /kr/ tends to be transliterated as <ခရ-> as the following examples show.

<CHRIST> → <ခရစ်> [M]

<CHROMIUM> → <ခရိုမီယမ်> [W]

When <CH> just stands for /k/, /k/ to <ခ-> may not be triggered,

<CINCHONA> → <စင်ကိုနာ> [M]

nor likely be triggered in <CHL> for /kl/.

<CHLORINE> → <ကလိုရင်း> [W]

A similar and strong mapping is <TR> to <ထရ->, where the aspirated <ထ> rather than common <တ> is used,

<ELECTRON> → <အီလက်ထရွန်> [M]

<TRANSISTOR> → <ထရန်စစ္စတာ> [M]

or even irregular spellings as <TR> to <တြ->.

<GEOMETRY> → <ဂျီသြမေတြိ> [M]

<DISTRICT> → <ဒိစတြိတ်> [M]

<BR> to <ဗြ-> (regularly, <ဘရ->) can also be observed.

<BRITAIN> → <ဗြိတိန်> [W]

## 4.5    Null Onset and Hiatus

When the onset of a syllable is absent, <အ-> (for native /ʔ/) will be used as a common placeholder. This can appear at the beginning of a word, or in a hiatus within one word. The following example contains both.

<IODINE> → <အိုင်အိုဒင်း> [M]

When the hiatus begins with /i/, <ယ-> (or <ရ->) instead of <အ-> is usually used.

<UNION> → <ယူနီယန်> [M]

Notice /ju/ to <ယူ> is used at word beginning to avoid the combination of <အျ>. <+ျူ> for /ju/ is common after a general onset.

<MERCURY> → <မာကျူရီ> [M]

Sometimes <ဝ-> will be inserted after /u/.

<LOUISIANA> → <လူဝီစီယားနားး> [W]

In some stable borrowed words, independent vowel letters are used. The following examples use such irregular spellings at word beginning.

<APRIL> → <ဧပြီ> [M]

<AUGUST> → <သြဂုတ်> [M]

For triphthongs (by some analysis), the syllable may be re-segmented with semi-vowel insertion.

<POWER> → <ပါဝါ> [M]

In this example, /aʊɪ/ is analyzed as /a.wɹ/. As another example,

<WIRE> → <ဝိုင်ယာ> [M]

here /aɪr/ is analyzed as /aɪ.ɹ/ with a nasalization (as there is no bare /aɪ/ rhyme in Burmese) and an inserted <ယ->.

## 4.6 Nucleus

The high tone and low tone of Burmese rhymes are largely exchangeable in the transliteration. The low tone rhymes are used by default in the following descriptions unless the high tone counterparts have further special uses.

As Burmese has a typical seven vowel inventory, to phonetically transcribe most English vowels in open syllables is not difficult, such as /i/ to <+ီ>, /u/ to <+ူ>, /eɪ/ to <+ေ>, /oʊ/ to <+ို>, /ɔ/ to <+ေ႟ာ>, and /ɑ/ to <+ာ>. These mappings are usually extended to some strong grapheme-to-grapheme mappings regardless of the phonemes, especially for non-native English words, such as <A> to <+ာ>, <E>/<EE>/<I>/<Y> to <+ီ>, <AY> to <+ေ>, <O> to <+ို>, <U>/<OO> to <+ူ>, and <AU>/<AW> to <+ေ႟ာ>.

/ɪ/ to <+ိ> and /ɛ/ to <+က်> are very common for lax vowels in closed syllables.

The diphthongs in Burmese always have a glottal ending or nasalization to form a rhyme. The nasalized rhymes are commonly used for English diphthongs in most cases such as /aɪ/ to <+ိုင်>

and /aʊ/ to <+ေ႟ာင်>; the corresponding glottal rhymes are more used before obstruents. There is no /ɔɪ/ in Burmese, which is commonly treated as /waɪ/ by <+ဝိုင်> in transliteration.

Reduced vowels (schwa) in English may be represented by the inherit vowel or based on the graphemes by surface spellings. Silent <E> at word/morpheme ending, and syllabized <ER> an <EL> may be stressed by a high tone.

## 4.7 Sonorant Coda

The nasal codas of /n/, /m/, /ŋ/ are generally mapped to the nasalized rhymes in Burmese. The articulation positions are not distinguished. The nucleus of Burmese nasalized rhymes can be /a/, /ɪ/, /ʊ/, /aɪ/, /aʊ/, /eɪ/, and /oʊ/, where the mid monophthongs are missing, for which rhymes of /aN/, /ɪN/, and /ʊN/ are imprecisely used.

When there are doubled nasal letters in spelling, the previous vowels may be represented as a nasalized rhyme no matter whether the consonant is geminate.

<COMMUNIST> → <ကွန်မြူနစ်> [M]

<FLANNEL> → <ဖလန်နယ်> [M]

Anyway, it is not always so.

<COMMISSION> → <ကော်မရှင်> [M]

<PENNY> → <ပဲနီ> [M]

Irregularly, the letters may be stacked for nasalized rhyme.

<SUMMONS> → <သမ္မန်> [M]

Sometimes the preceding vowels is by a nasalized rhyme even in an open syllable.

<CAMERA> → <ကင်မရာ> [M]

Generally, vowel letters followed by <R> will be transliterated by <+ာ>.

<HARMONICA> → <ဟာမိုနီကာ> [M]

<MERCURY> → <မာကျူရီ> [M]

<VIRGINIA> → <ဗာဂျီးနီးယား> [M]

<JOURNAL> → <ဂျာနယ်> [M]

<TURPENTINE> → <တာပင်တိုင်> [M]

An exception is <OR>, which may be kept as <+ဉ်ာ> or lowered to <+ေဟာင်>, say, the rounded feature is reserved.

<FOREMAN> → <ဖိုမင်> [M]

<TORPEDO> → <တော်ပီဒိုု> [M]

Unstressed suffixes such as <AR>, <ER>, <OR>, and <URE> are also transliterated by <+တာ>.

<DOLLAR> → <ဒေါ်လာ> [M]

<DRIVER> → <ဒရိုင်ဘာ> [M]

<EDITOR> → <အယ်ဒီတာ> [M]

<TINCTURE> → <တင်ချာ> [M]

Syllabized <R> is generally treated as <ER>.

<METER>/<METRE> → <မီတာ> [M]

Vowel letters followed by <L> will be generally transliterated by <+ယ်>, or <+ယ်လ်>. This is especially for <AL> or <EL>.

<ALBUM> → <အယ်လဘမ်> [M]

<ROYAL> → <ရွိုင်ရယ်> [M]

<HOTEL> → <ဟိုတယ်> [M]

<PARCEL> → <ပါဆယ်> [M]

In <IL> and <UL>, the <L> may be dropped.

<COUNCIL> → <ကောင်စီ> [M]

<PULLEY> → <ပူလီ> [M]

In <OL>, the <L> may be dropped an the <O> is lowered to <+ေဟာင်> or kept as <+ဉ်ာ>.

<COLLAR> → <ကော်လာ> [M]

<VOLT> → <ဗို့> [M]

Syllabized <L> is generally treated as <EL>.

<BICYCLE> → <ဘိုင်စကယ်> [M]

When there are multiple sonorant consonants in a coda, the nasal consonants are usually kept by a nasalized rhyme and <R> and <L> are dropped or reduced into a high tone.

<HORN> → <ဟွန်း> [M]

Sometimes, it will be analyzed into multiple syllables for a complex coda with multiple sonorant consonants. This is like the case of triphthongs.

<FILM> → <ဖလင်> [M]

## 4.8 Obstruent Coda

A syllable with a coda having low sonority is generally transliterated into a rhyme with a glottal ending or a creaky tone; the obstruents of low sonority in coda are then dropped. The nucleus of Burmese glottal ended rhymes can be /a/, /ɛ/, /ı/, /ʊ/, /aı/, /aʊ/, /eı/, and /oʊ/, where the mid monophthong /ɛ/ is absent in nasalized rhymes but present here (while /ɔ/ is still absent).

<SALAD> → <ဆာလတ်> [M]

When there are double letters (including <CK>) within the word, the glottal ended rhymes are also used, no matter whether there is a geminate.

<BATTERY> → <ဘက်ထရီ> [M]

<RACKET> → <ရက်ကက်> [M]

As the nasalized rhymes cannot take a glottal ending, the creaky tone is generally added.

<CONFERENCE> → <ကွန်ဖရင့်> [M]

<POINT> → <ပွိုင့်> [M]

<SECOND> → <စက္ကန့်> [M]

The creaky tone also tends to be used in <+ေဟာဉ့ာ> and <+ဉ်ဉ့ာ> for <O>.

<LOCKET> → <လော့ကက်> [M]

<POSTCARD> → <ပို့စ်ကတ်> [M]

This is even for a plain open syllable.

<NOTICE> → <နိုတ်စ်> [M]

In more modern transliteration instances, the obstruents in a coda may be further denotated by inherent vowel suppressed letters which is non-native to Burmese orthography. The following are some instances observed on Wikipedia.

Although <+က်>, <+စ်>, <+တ်>, and <+ပ်> represent regular Burmese rhymes, they can be further used after a normal rhyme.

<CYRILLIC> → <ဆရရ်လစ်က်> [W]

<BUS> → <ဘတ်စ်> [W]

<KHIGHT> → <နိုက်တ်> [W]

<PHILIP> → <ဖိလစ်ပ်> [W]

The use of <ထ်> and <ခ်> can also be observed.

<ERNST> → <အန့်ထ်> [W]

<MARK> → <မာ့ခ်> [W]

<ဘ်>, <ဒ်>, and <ဂ်> are respectively for the voiced stops /b/, /d/, /g/ in coda.

<WEB> → <ဝက်ဘ်> [W]

<OXFORD> → <အောက်စ်ဖိုဒ်> [W]

<PRAGUE> → <ပရက်ဂ်> [W]

For /tʃ/ and /dʒ/ in coda, <ချ်> is commonly used; <ဂျ်> is less common.

<DUTCH> → <ဒတ်ချ်> [W]

<CAMBRIDGE> → <ကိန်းဘရစ်ချ်> [W]

<GEORGE> → <ဂျော့ဂျ်> [W]

<ဖ်> is reasonably for /f/ in coda.

<KAFKA> → <ကပ်ဖ်ကာ> [W]

<ဖ်> may appear in Russian names.

<CHEKHOV> → <ချက်ကော့ဖ်> [W]

Beside <စ်>, <ဇ်> and <ရှ်> are also used for sibilants in coda.

<HERTZ> → <ဟာ့တ်ဇ်> [W]

<PUSHKIN> → <ပုရှ်ကင်> [W]

<သ်> can be used for coda of <TH>. In the following instance, a doubled <ဒ်သ်> is also used.

<WORDSWORTH> → <ဝါ့ဒ်သ်ဝါ့သ်> [W]

In quite recent words, the transliteration tends to be mechanical, thus these extra letters are more often used.

<FACEBOOK> → <ဖေ့စ်ဘွတ်ခ်> [W]

<GITHUB> → <ဂစ်တ်ဟပ်�’> [W]

<YOUTUBE> → <ယူကျူ(ဘ်)> [W]

In some cases, the coda will be fully transliterated without depressing the inherent vowel.

<ABDUL> → <အာ�‌ဘဒူ> [W]

## 5 Conclusion and Future Work

In this manuscript, we provide a general investigation on the transliteration of foreign words in Burmese. These materials are expected to be organized in a more normalized form for a more prescriptive guidelines as our future work.

A Sub-syllabic Segmentation Scheme for Statistical Solutions. In Proc. of PACLING, CCIS 781, pp. 191—202.

Aye Myat Mon, C. Ding, H. Kaing, Khin Mar Soe, M. Utiyama, and E. Sumita (2020), A Myanmar (Burmese)-English Named Entity Transliteration Dictionary. In Proc. of LREC, pp. 2973—2976.

## Appendix

### Transliteration of Chinese into Burmese

The analysis is based on around 150 instances of place names (provinces, cities, etc.) in China collected from the Burmese Wikipedia. The Latin graphemes are based on Pinyin.

This is a quite general analysis to map Pinyin to Burmese spellings, where irregular but fixed transliteration instances are not covered and left to the investigation in future.

### 1 Initials

The correspondence is clear as follows, <B> to <ပ->, <P> to <ဖ->, <M> to <မ->, <F> to <ဖ->, <D> to <တ-> or <ဒ->, <T> to <ထ->, <N> to <န->, <L> to <လ->, <G> to <က->, <K> to <ခ->, <H> to <ဟ->, <J> to <ကျ->, <Q> to <ချ->, <X> to <ရှ->, <ZH> to <ကျ->, <CH> to <ချ->, <SH> to <ရှ->, <R> to <ရ->, <Z> to <ဇ->, <C> to <ဆ->, <S> to <စ->, <Y> to <ယ->, <W> to <ဝ->, and <> (/ʔ/) to <အ->.

Chinese and Burmese both have the contrast of unaspirated and aspirated stops. The <ထ-> and <ဆ-> not often used for English are thus used here. For <D>, the voiced <ဒ-> can be observed sometimes. The <P> and <F> cannot be distinguished that both are represented by <ဖ->. <J>/<ZH>, <Q>/<CH>, <X>/<SH> are not distinguished. <R> is not common but generally observed by <ရ->. The unaspirated affricate <Z> and aspirated affricate <C> are by voiced <ဇ-> and aspirated <ဆ->, respectively. The <Y> and <W> can be a placeholder for an empty initial /ʔ/ (e.g., <WU>), or a medial with the empty initial (e.g., <WA>). The transliteration seems commonly based on the Pinyin spellings, such as <ဝု> (rather than <ဝှု>) for <WU>.

### 2 Finals

The three medials of Chinese /j/, /w/, and /ɥ/ are transliterated by <+ျ->, <+ွ->, and <+ျွ->, respectively. The <+ျ-> will be contracted with the initials with it, say, <ကျ-> and <ချ->, and will be dropped with <ရှ->. With the initial of /ʔ/, they generally become <ယ->, <ဝ->, and <ယွ->, respectively. Non-native combinations such as <ကျ-> for /tj/ can be observed. Sometimes it becomes <တယ-> to adapt the native phonotactics.

The finals of open syllables and falling diphthongs in Chinese can be mapped to Burmese relatively clearly on phonemes, specifically, <I> to /i/, <U> to /u/, <A> to /a/, <E> to /ɛ/, <UO> to /ɔ/, <AI> to /aɪɴ/, <EI> to /e/, <AO> to /aʊɴ/, and <O>/<UO> to /o/.

Like the medials, <Ü> is treated as /ju/. The <Ê> and <O> are marginal in Chinese and <E> and <UO> are commonly mapped to the two open-mid vowels in Burmese.[3] <IE> and <ÜE> are further treated as /jɛ/ and /ɥɛ/ respectively. <EI> and <OU> are mapped to the two close-mid vowels, and <AI> and <AU> are mapped to two nasalized diphthongs in Burmese. <IA>, <UA>, <UAI>, and <IAO> are treated as /ja/, /wa/, /waɪɴ/, and /jaʊɴ/, respectively. <UI> is /we/ and <IU> is /jo/.

The <I> after <Z>, <C>, and <S> may be mapped to /ɿʔ/ and represented by <+ စ်>, but they are generally treated as a plain /i/. The <ER> is represented by <+ အာရ်>.

The finals with nasal ends of "N" and "NG" in Chinese cannot be distinguished exactly in Burmese. Only the nucleus can be distinguished to a certain extent. <AN>/<ANG> and <IN>/<ING> by /aɴ/ and /ɪɴ/ are relatively stable. <EN>/<ENG> may be merged with <IN>/<ING> or mapped to

---

[3] In <UO> the medial /w/ is omitted and no necessary to insert a <+ွ->.

/eɪN/. <ONG> mapped to /oʊN/ is stable.[4] <IAN>, <IANG>, and <IONG> are reasonably treated as /jaN/, /jaN/, and /joʊN/, respectively. The medial /w/ for <UAN>, <UN>, and <UANG> are represented by the insertion of <+ွ-> before <AN>, <EN>, and <ANG>, even this may round the rhymes to /ʊN/ (Okano, 2007). Similarly, the medial /ɥ/ for <ÜAN> and <ÜN> is thus by the insertion of <+ျွ-> before <AN> and <EN>.

As to the issues of orthography, the <+ၞ်> and <+ဉ်> are both used, while <+ဉ်> may be more used for "NG". <ONG> to <+ၟ်> is common.

The nasal ending may be dropped in special cases when the transliteration is not by characters but by words. Here are examples where the <-N> in the first syllables are dropped.

<YUNNAN> → <ယွန်နန်> [W]

<KUNMING> → <ကွန်မင်း> [W]

# 3  Tones

The Burmese high tone for Chinese T1 (55) and creaky tone for T4 (51) are generally stable. The Burmese low tone for Chinese T3 (214) is also common. The correspondent of Chinese T2 (35) may be the low tone, but sometimes the high tone is used. This is unstable, as there is no rising tone in Burmese.

---

[4] It has not been attested yet whether <WENG> is treated as     <ONG> in temporary data.